\documentclass{article}

     \usepackage[preprint]{neurips_2019}

\usepackage[utf8]{inputenc} 
\usepackage[T1]{fontenc}    
\usepackage{hyperref}       
\usepackage{url}            
\usepackage{booktabs}       
\usepackage{amsfonts}       
\usepackage{nicefrac}       
\usepackage{microtype}      
\usepackage{amsmath}
\usepackage{graphicx}
\usepackage{subfig}

\usepackage{times}
\usepackage[utf8]{inputenc} 
\usepackage[T1]{fontenc}    
\usepackage{hyperref}       
\usepackage{url}            
\usepackage{booktabs}       
\usepackage{amsfonts,amsthm}       
\usepackage{nicefrac}       
\usepackage{microtype}      
\usepackage{amsmath}
\usepackage{dirtytalk}
\usepackage{tikz}

\usepackage{amsmath}
\usetikzlibrary{bayesnet}
\usetikzlibrary{positioning}

\newcommand{\params}{\boldsymbol{\theta}}

\definecolor{green}{HTML}{60ba46}
\definecolor{blue}{HTML}{1749b3}
\definecolor{orange}{HTML}{fa9d00}

\newcommand{\G}{\textcolor{blue}{G}}
\newcommand{\Gin}{\textcolor{green}{G_{\overline{T}}}}
\newcommand{\Gout}{\textcolor{orange}{G_{\underline{T}}}}

\title{Causal inference with Bayes rule}

\author{%
Finnian Lattimore\\
The Gradient Institute, Paris\\
\And
David Rohde\\
Criteo AI Lab, Paris\\
}

\newcommand{\ci}{\perp\!\!\!\perp}

\newcommand{\eqn}[1]{\begin{align}#1\end{align}}
\newcommand{\eq}[1]{\begin{align*}#1\end{align*}}

\renewcommand{\P}[1]{\mathbb{P}(#1)}

\newcommand{\bt}{\boldsymbol{t}}

\begin{document}

\maketitle

\begin{abstract}
  The concept of causality has a controversial history. The question of
  whether it is possible to represent and address causal problems with
  probability theory, or if fundamentally new mathematics such as the do-calculus is required
  has been hotly debated,  In this paper
  we demonstrate that, while it is critical to explicitly model our
  assumptions on the impact of intervening in a system, provided we do
  so, estimating causal effects can be done entirely within the standard
  Bayesian paradigm. The invariance assumptions underlying causal
  graphical models can be encoded in ordinary Probabilistic graphical models,
  allowing causal estimation with Bayesian statistics, equivalent to the
  do-calculus. 
\end{abstract}




\section{Introduction}

Difficult causal questions, such as \emph{`does eating meat cause cancer'} or \emph{`would increasing the minimum wage lead to a fall in employment'} are fundamental to decisions around how our society is structured and our understanding of the world. The development of Causal Graphical Models (CGMs) and the do-calculus \cite{pearl1995causal,pearl2009} has given us an extremely rich and powerful framework with which to formalise and approach such questions. This framework is presented as fundamentally extra-statistical - Pearl has argued forcefully that (Bayesian) probability theory alone is not sufficient for solving causal problems \cite{pearl2001bayesianism}. 

The notion that causality fundamentally requires new mathematics and that causal questions cannot be solved within existing paradigms for probabilistic inference has led to extensive controversy and debate, eg \cite{gelman2009blog,gelman2019blog}. This debate has been particularly intense between proponents of causal modelling and Bayesian modellers, perhaps not surprisingly, since the Bayesian approach to combining assumptions with data is typically presented as sufficiently general to tackle \emph{any} probabilistic inference problem (although computational constraints may make it impractical).

In this paper, we demonstrate how the assumptions encoded by causal graphical models can be represented with a probabilistic graphical model (PGM). The advantage of doing so is mostly conceptual: it allows Bayesian practitioners to represent and reason about the modelling assumptions required for causal inference in a framework with which they are familiar. However, there may also be practical benefits in cases where causal queries are not identifiable via the do-calculus. In such cases, it is fundamentally impossible to infer the exact outcome of an intervention, even given infinite pre-interventional data without additional assumptions. Modelling such problems within a standard Bayesian inference setting allows us to leverage a vast body of existing research on combining assumptions with data to obtain finite sample estimates for distributions of interest. While the posterior distribution will always remain sensitive to the prior (unless we add assumptions about the functional form of the relationships between variables) we may still obtain useful bounds. The disadvantage of modelling causal questions explicitly as a single PGM is that it is more cumbersome and computationally expensive (unless we use the machinery of the do-calculus to identify appropriate re-parameterisations). 

\subsection{Representing a Causal Problem with a Probabilistic graphical model}
In the following sections we show how a causal query can be represented with a PGM and how to do causal inference via this approach. For the necessary background on probabilistic and causal graphical models, we refer readers to the appendix. 

To represent an intervention with an ordinary Probabilistic graphical model, we must explicitly model the pre and post intervention systems and the relationship between them. Algorithm 1 constructs a probabilistic graphical model for a specific intervention in a causal graphical model. 
\paragraph*{Algorithm 1: CausalBayesConstruct}\label{Alg:causebayesconstruct}~\\ \emph{Input}: Causal graph $G$ and intervention $do(T=t)$. ~\\\emph{Output}: Probabilistic graphical model representing this intervention
\begin{enumerate}
\item Draw the original causal graph $G$ inside a plate indexed from $1, ... M$ to represent the data generating process.
\item For each variable $V\in G$, parameterize $P(V|parents(V))$ by adding a parameter $\theta_V$ with a link into $V$. 
\item Draw the graph after the intervention by setting $T=t$ and
  removing all links into it. 
Rename each of the variables to distinguish them from the variables in the original graph, e.g. $X$ becomes $X^*$.
\item Connect the two graphs linking $\theta_V$ to the corresponding variable $V^*$ in the post-interventional graph, for each $V$ excluding $T$.


\end{enumerate}

\begin{figure}
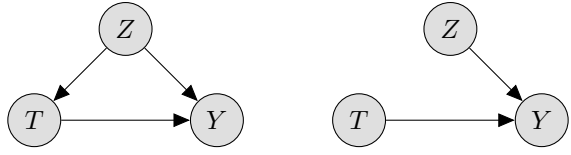

  \centering
  \tikz{ %
    \node[obs] (Zm) {$Z$} ; %
    \node[obs, below left=of Zm] (Xm) {$T$} ; %
    \node[obs, below right=of Zm] (Ym) {$Y$} ; %
    \edge {Zm} {Xm} ; %
    \edge {Zm} {Ym} ; %
    \edge {Xm} {Ym} ; %
  } \hspace{1cm}
  \tikz{ %
    \node[obs] (Zm) {$Z$} ; %
    \node[obs, below left=of Zm] (Xm) {$T$} ; %
    \node[obs, below right=of Zm] (Ym) {$Y$} ; %
    \edge {Zm} {Ym} ; %
    \edge {Xm} {Ym} ; %
  }
  \caption{A CGM of Case 1: Left observational, Right: mutilated}
  \label{fig:cgm1}  
\end{figure}

\begin{figure}
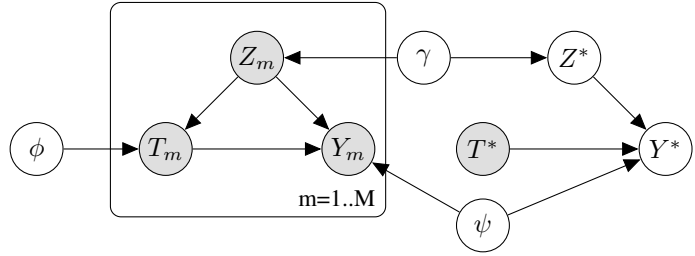

  \centering
  \tikz{ %
    \node[obs] (Zm) {$Z_m$} ; %
    \node[obs, below left=of Zm] (Xm) {$T_m$} ; %
    \node[obs, below right=of Zm] (Ym) {$Y_m$} ; %
    \edge {Zm} {Xm} ; %
    \edge {Zm} {Ym} ; %
    \edge {Xm} {Ym} ; %
    \plate[inner sep=0.25cm, xshift=-0.12cm, yshift=0.12cm] {plate1} {(Xm) (Zm) (Ym)} {m=1..M};
    \node[latent, right=3.5cm of Zm] (Zn) {$Z^*$} ; %
    \node[obs, below left=of Zn] (Xn) {$T^*$} ; %
    \node[latent, below right=of Zn] (Yn) {$Y^*$} ; %
    \edge {Zn} {Yn} ; %
    \edge {Xn} {Yn} ; %
    \node[latent, right=1.5cm of Zm] (gamma) {$\gamma$};
    \node[latent, left=of Xm] (phi) {$\phi$};
    \node[latent, below=.3cm of Xn] (theta) {$\psi$};
    \edge {gamma} {Zm,Zn};
    \edge {phi} {Xm};
    \edge {theta} {Ym,Yn};    
  }
  \caption{A PGM of Case 1}
  \label{fig:pgm1}  
\end{figure}

A PGM constructed with Algorithm 1 represents exactly the same assumptions about a specific intervention as the corresponding CGM, see Figures~\ref{fig:cgm1} and \ref{fig:pgm1} for an example. We have just explicitly created a joint model over the system pre and post-intervention, which allows the direct application of standard statistical inference, rather than requiring additional notation and operations that map from one to the other - as the do-calculus does. The Bayesian model is specified by the parameterization of the conditional distribution of variables given their parents, and priors may be placed on the parameters $\params$. The fact that the parameters are shared for all pairs of variables $(V,V^*)$ excluding $T$, captures the assumption that all that is changed by the intervention is the way $T$ takes its value - the conditional distributions for all other variables given their parents are invariant. 

Despite its simplicity we are unaware of a direct statement of Algorithm 1, it is related to twin networks \cite{pearl2009} and augmented directed acyclic graphs \cite{dawid2015statistical} but is distinct from both.

\subsection{Causal Inference with Probabilistic graphical models}
The result of Algorithm 1 is a Probabilistic graphical model on which we can do inference with standard probability theory rather than the do-calculus, and which has properties such as arrow reversal (by the use of Bayes rule). To infer causal effects we compute a predictive distribution for the quantity of interest in the post-intervention graph using Bayes rule, integrating out all parameters, latent variables and any observed variables that are not of interest, for each setting of the treatment $T=t^*$. 


\newcommand{\Dpost}{\boldsymbol{v^*}}
\newcommand{\Dpre}{\boldsymbol{v}}

To make this procedure clearer, let $\boldsymbol{V}$ be the set of variables in the original causal graph $\G$, excluding the variable we intervene on, $T$, and $\boldsymbol{V^*}$ be the corresponding variables in the post-interventional graph. We have: $\params$: the set of model parameters, $\Dpre$: a matrix of the $M$ observations of variables $\boldsymbol{V}$, $(\boldsymbol{v}_1,...,\boldsymbol{v}_M)$ collected pre-intervention,
$\bt$: a vector of the $M$ observed values of the treatment variable $T$, $t_1,..t_M$, 
$\Dpost$: The variables of the system post-intervention,
$t^*$: the value that the intervened on variable $T$ is set to,
$Y^*\in {\Dpost}$: the variable of interest post-intervention.


The goal is to infer the value of the unobserved post-interventional
distribution over $\Dpost$, given the observed data and $(\Dpre,\bt)$
and a selected treatment $t^*$. By construction, conditional on the
parameters $\params$, the post-interventional variables $\Dpost$ are
independent of data collected pre-intervention $(\Dpre,\bt)$. The
value of the intervention $t^*$ is set exogenously\footnote{Also $t^*$
  has no marginal distribution - it is a constant set by the
  intervention} - so is independent of both $\params$ and
$(\Dpre,\bt)$. This ensures joint distribution over
$(\Dpre,\bt,\Dpost,\params)$ factorize into three terms: a prior over
the parameters $\P{\params}$, the likelihood for the original system
$\P{\Dpre,\bt|\params}$, and a predictive distribution for the
post-interventional variables given parameters and intervention
$\P{\Dpost|\params,t^*}$:

\eq{
\P{\Dpre,\bt,\Dpost,\params|t^*} = \P{\params}\P{\Dpre,\bt|\params}\P{\Dpost|\params,t^*}
}


We then marginalize out $\params$,
\eqn{
\P{\Dpre,\bt,\Dpost|t^*} = \int_{\params}\P{\params}\P{\Dpre,\bt|\params}\P{\Dpost|\params,t^*}d\params
}
and condition on the observed data $(\Dpre,\bt)$,
\eqn{
\P{\Dpost|\Dpre,\bt,t^*} &= \frac{\P{\Dpre,\bt,\Dpost|t^*}}{\P{\Dpre,\bt|t^*}} \nonumber \\
&= \int_{\params}\frac{\P{\params}\P{\Dpre,\bt|\params}}
{\P{\Dpre,\bt}}\P{\Dpost|\params,t^*}d\params \nonumber \\
&= \int_{\params}\P{\params|\Dpre,\bt}\P{\Dpost|\params,t^*}d\params.
\label{eqn:post_given_observed_general}
}
Finally, if the goal is to infer mean treatment effects\footnote{We could also compute conditional treatment effects by first conditioning on selected variables in $\Dpost$.} on a specific variable post-intervention $Y^*$, we can marginalize out the remaining variables in $\boldsymbol{V^*}$, 
\eqn{
\P{Y^*|\Dpre,\bt,t^*}
&= \int_{\params}\P{\params|\Dpre,\bt}\sum_{\boldsymbol{V^*}\backslash Y^*}\P{\Dpost|\params,t^*}d\params.
\label{eqn:post_given_observed_general}
}

If there are no latent variables in $\G$, assuming positive density over the domain of $(\Dpre,\bt)$ and a well defined prior $\P{\params}$, the likelihood  $\P{\Dpre,\bt|\params}$ will dominate, and the posterior over the parameters $\P{\params|\Dpre,\bt}$ will become independent of the prior at the infinite data limit. The term $\P{\Dpost|\params,t^*}$ can be expanded into a product of terms of the form $\P{V^*|parents(V^*),\params}$ following the factorization implied by the post-interventional graph. From step (3) of Algorithm 1 each of these terms are equal to the corresponding terms $\P{V|parents(V),\params}$, giving results equivalent to Pearl's truncated product formula \cite{pearl2009}. \citet{lattimore2019replacing} demonstrate the equivalence of this approach with the do-calculus on a number of worked examples.



\section{Conclusion}

The paper shows that it is possible to arrive at the same solution for
causal problems using both the do-calculus and Bayesian theory, the
key insight required for the Bayesian formulation is that the probabilistic
graphical model must jointly model both the pre-intervention and post
intervention worlds. Our conclusion is similar to that of \cite{lindley1981role}, however we provide an explicit mechanism by which we can encode the assumptions implied by a causal graphical model, formalising the notion of exchangability in this context.

\bibliography{literature}

\section{Appendix}
\subsection{Background on Probabilistic and Causal graphical models}
Probabilistic graphical models (PGMs) combine graph theory with
probability theory in order to develop new algorithms and to present
models in an intuitive framework \cite{jordan2004graphical}. A Probabilistic graphical model is a directed acyclic graph over variables, which represents how the joint distribution over these variables may be factorized. In particular, any \emph{missing} edge in the graph must correspond to a conditional independence relation in the joint distribution. There are multiple valid Probabilistic graphical model representations for a given joint distribution. For example, any joint distribution over two variables $(X,Y)$ may be represented by both $X \rightarrow Y$ or $X \leftarrow Y$.



A causal graphical model (CGM) is a Probabilistic graphical model, with the additional assumption that a link $X \rightarrow Y$ means $X$ causes $Y$. Think of the data generating process for a CGM as sampling data first for the exogenous variables (those with no parents in the graph), and then in subsequent steps sampling values for the children of previously sampled nodes. An atomic intervention in such a system that sets the value of a specific variable $T$ to a fixed constant corresponds to removing all links into $T$ - as it is now set exogenously, rather than determined by its previous causes. It is assumed that everything else in the system remains unchanged,  in particular the functions or conditional distributions that determine the value of a variable given its parents in the graph. In this way, a CGM encodes more than the factorization (or conditional independence structure) of the joint distribution over its variables; It additionally specifies how the system responds to atomic interventions. 

A CGM describes how the structure of a system is modified by an intervention. However, answering causal queries such as \textit{"what would the distribution of cancer look like if we were able to prevent smoking?"} requires inference about the distributions of variables in the post-interventional system. The do-notation is a short-hand for describing the distribution of variables post-intervention and the do-calculus is a set of rules for identifying which (conditional) distributions are equivalent pre and post-intervention. If it is possible to derive an expression for the desired post-interventional distribution purely in terms of the joint distribution over the original system via the do-calculus then the causal query is identifiable, meaning assuming positive density and infinite data we obtain a point estimate for it. 

Here we present the do-calculus in a simplified form that applies to interventions on single variables  \cite{pearl1995causal,pearl2009,peters2017elements}. 

\paragraph*{The do-calculus} Let $G$ be a CGM, $\Gin$ represent $G$ post-intervention (i.e with all links into $T$ removed) and $G_{\underline{T}}$ represent $G$ with all links \emph{out of} $T$ removed. Let $do(t)$ represent intervening to set a single variable $T$ to $t$,

\subparagraph{Rule 1:} $\P{y|do(t),z,w}=\P{y|do(t),z}$ if $Y\ci W|(Z,T)$ in $\Gin$
\subparagraph{Rule 2:} $\P{y|do(t),z}=\P{y|t,z}$ if $Y\ci T|Z$ in $\Gout$
\subparagraph{Rule 3:} $\P{y|do(t),z}=\P{y|z}$ if $Y\ci T|Z$ in $\Gin$, and $Z$ is not a decedent of $T$.

\end{document}